\documentclass{article}
\usepackage{graphicx}
\usepackage[colorlinks=true,urlcolor=blue]{hyperref}
\usepackage{authblk}
\usepackage[numbers,sort&compress]{natbib} 
\usepackage{caption}
\usepackage{float}
\usepackage{caption}
\usepackage{appendix}

\title{Zebra-Llama: A Context-Aware Large Language Model for Democratizing Rare Disease Knowledge}

\author[1\thanks{Corresponding author: karthi.soman@gmail.com or karthik.soman@ucsf.edu}]{Karthik Soman}
\author[2]{Andrew Langdon}
\author[1]{Catalina Villouta}
\author[3]{Chinmay Agrawal}
\author[2]{Lashaw Salta}
\author[1]{Braian Peetoom}
\author[1,4]{Gianmarco Bellucci}
\author[5]{Orion J Buske}
\affil[1]{Dept. of Neurology, University of California San Francisco, California, USA}
\affil[2]{Independent Researcher}
\affil[3]{Plix AI, California, USA}
\affil[4]{Dept. of Neurosciences, Mental Health and Sensory Organs, Sapienza University of Rome, Italy}
\affil[5]{PhenoTips, Toronto, Canada}

\date{}

\begin{document}

\maketitle

\begin{abstract}
Rare diseases present unique challenges in healthcare, often suffering from delayed diagnosis and fragmented information landscapes. The scarcity of reliable knowledge in these conditions poses a distinct challenge for Large Language Models (LLMs) in supporting clinical management and delivering precise patient information—underscoring the need for focused training on these 'zebra' cases. We present Zebra-Llama, a specialized context-aware language model with high precision Retrieval-Augmented Generation (RAG) capability, focusing on Ehlers-Danlos Syndrome (EDS) as our case study. EDS, affecting 1 in 5,000 individuals, exemplifies the complexities of rare diseases with its diverse symptoms, multiple subtypes, and evolving diagnostic criteria. By implementing a novel context-aware fine-tuning methodology trained on questions derived from medical literature, patient experiences, and clinical resources, along with expertly curated responses, Zebra-Llama demonstrates unprecedented capabilities in handling EDS-related queries. On a test set of real-world questions collected from EDS patients and clinicians, medical experts evaluated the responses generated by both models, revealing Zebra-Llama's substantial improvements over base model (Llama-3.1-8B-Instruct) in thoroughness (77.5\% vs. 70.1\%), accuracy (83.0\% vs. 78.8\%), clarity (74.7\% vs. 72.0\%) and citation reliability (70.6\% vs. 52.3\%). Released as an open-source resource, Zebra-Llama not only provides more accessible and reliable EDS information but also establishes a framework for developing specialized AI solutions for other rare conditions. This work represents a crucial step towards democratizing expert-level knowledge in rare disease management, potentially transforming how healthcare providers and patients navigate the complex landscape of rare diseases.
\end{abstract}

\subsection*{Model Access}

\textbf{Model weights:} \\
\url{https://huggingface.co/zebraLLAMA/zebra-Llama-v0.2}
\newline
\newline
\textbf{Custom built RAG API for rare diseases (focused on EDS):}\\
\textbullet~Base URL: \url{https://zebra-llama-rag.onrender.com} \\
\textbullet~Endpoint: /search
\newline
\newline
\textbf{Jupyter Notebook Demo of Zebra-Llama:} \\
\url{https://github.com/karthiksoman/zebra-Llama/blob/main/code/notebook/zebra_llama_v0.2_demo.ipynb}
\newline
\newline
\textbf{Source code:} \\
\url{https://github.com/karthiksoman/zebra-llama}

\section{Introduction}

The advent of Large Language Models (LLMs) has ushered in a new era of artificial intelligence, with these models demonstrating unprecedented capabilities in natural language understanding and generation \cite{zhao2023survey}. Their potential applications span numerous fields, including healthcare, where they promise to revolutionize tasks ranging from information retrieval to clinical decision support. Recent efforts to adapt LLMs for medical applications broadly fall into three categories: specialized pre-training, domain-specific fine-tuning, and advanced prompting techniques. Notable examples include BioGPT, pre-trained on PubMed abstracts \cite{luo2022biogpt}, and MedAlpaca, fine-tuned on specialized medical datasets \cite{han2023medalpaca}. Advanced prompting strategies, exemplified by MedPaLM \cite{singhal2023large} and Medprompt \cite{nori2023can}, have achieved results comparable to human experts on medical licensing exams. While these approaches have yielded promising results in various medical tasks, they often do not fill the gap for highly specialized medical domains, particularly in the context of rare diseases.

Rare diseases present unique challenges in the medical field, affecting only a small portion of the population yet often eluding timely diagnosis and effective treatment \cite{griggs2009clinical, stoller2018challenge}. The medical adage, "When you hear hoofbeats, think horses, not zebras," while pragmatic for common ailments, can inadvertently lead to overlooking rare conditions – the metaphorical "zebras" of medicine \cite{smith2000you}. This oversight can result in significant delays in diagnosis and treatment, profoundly impacting patient outcomes and quality of life. The application of LLMs to rare disease management presents a promising yet underexplored avenue. Unlike common medical conditions with abundant data, rare diseases suffer from a scarcity of information and expertise, making them particularly challenging for traditional LLM approaches. This scarcity underscores the need for specialized AI solutions that can effectively aggregate, contextualize, and disseminate the limited available information on rare diseases from various sources, potentially transforming the landscape of rare disease diagnosis and management.

Ehlers-Danlos Syndrome (EDS) epitomizes these challenges, serving as a prime example of the "zebras" in medicine that often go unrecognized. As a group of inherited connective tissue disorders, EDS manifests through a diverse array of symptoms, including joint hypermobility, skin hyperextensibility, and tissue fragility \cite{malfait2020ehlers}. With a combined prevalence estimated at 1 in 5,000 individuals worldwide, EDS significantly impacts patient lives while frequently evading timely diagnosis \cite{malfait2020ehlers}. The syndrome's heterogeneous nature, comprising at least 13 subtypes with overlapping yet distinct clinical presentations, further complicates the diagnostic process \cite{meester2017differences}. This complexity, coupled with the syndrome's relative rarity, often results in a prolonged diagnostic journey, misdiagnosis, and suboptimal management \cite{byers2017diagnosis}. Moreover, the rapid evolution of EDS research, with emerging genetic markers and updated clinical criteria, creates an ever-expanding knowledge base that is challenging to navigate \cite{caliogna2021biomarkers, tinkle2017hypermobile, malfait2010clinical}. EDS thus presents an ideal case study for exploring how specialized AI solutions, particularly advanced LLMs, could revolutionize rare disease management by effectively aggregating, contextualizing, and disseminating the limited yet crucial information available from various sources.

To address these challenges in EDS management and demonstrate the potential of specialized AI in rare disease contexts, we present Zebra-Llama, a novel context-aware language model tailored for EDS information management. Drawing from a diverse array of sources including medical literature, patient forums, and clinical resources, our model is designed to address the multifaceted nature of EDS concerns, ranging from clinical presentations to daily living challenges. Our approach significantly enhances the utility of Retrieval-Augmented Generation (RAG) in the EDS domain, as illustrated in Fig 1. While traditional base language models (Fig. 1A) may provide inaccurate or imprecisely cited information, and standard RAG implementations (Fig. 1B) often struggle with contextual relevance, Zebra-Llama (Fig. 1C) demonstrates superior performance in generating precise and accurately cited responses. This is achieved through our specialized context-aware fine-tuning methodology, which trains the model to effectively discern and utilize relevant information from retrieved contexts. Zebra-Llama not only improves the accuracy and relevance of generated answers but also enhances the model's ability to provide proper citations, a critical feature for trustworthy medical information systems. By focusing on EDS, our model exemplifies how tailored AI solutions can potentially transform rare disease information management, offering a scalable approach to aggregating, contextualizing, and disseminating specialized medical knowledge. Our results demonstrate significant improvements in thoroughness, accuracy, clarity, and citation reliability compared to base models, paving the way for more accessible and reliable rare disease knowledge dissemination. This work represents a crucial step towards leveraging advanced AI to address the "zebras" of medicine, potentially improving diagnosis, treatment, and overall care for individuals affected by EDS and, by extension, other rare conditions.

\begin{figure}[H]
    \centering
    \includegraphics[width=1\textwidth]{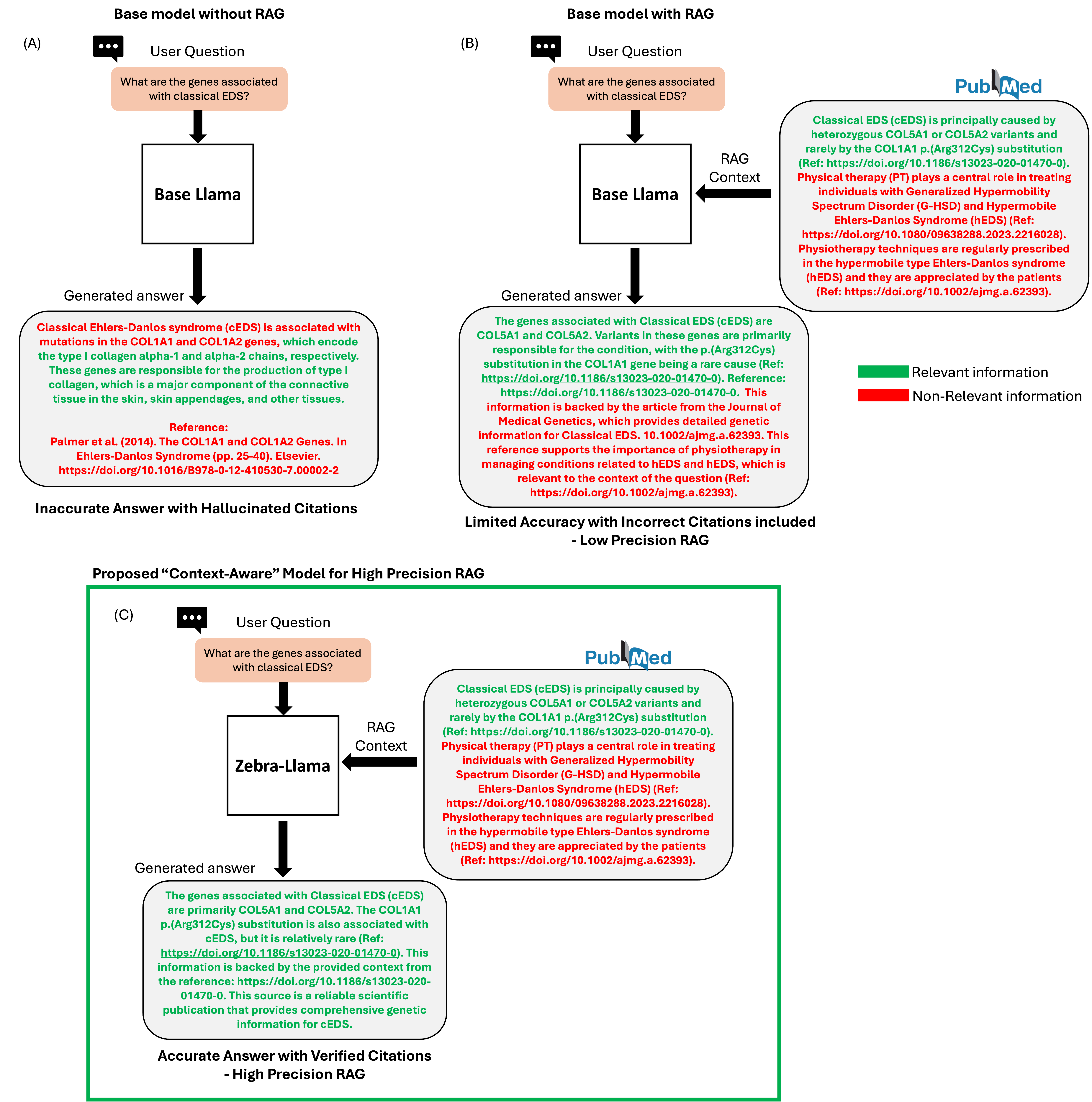}
    \caption{\textbf{Fig 1: Comparison of different approaches to EDS-related query handling.} (A) Base Llama model generating answers without RAG context, resulting in potentially inaccurate information and hallucinated citations. (B) Base Llama model with RAG implementation, showing imprecise utilization of retrieved context and inclusion of irrelevant information and citations. (C) Zebra-Llama model demonstrating enhanced context-aware RAG capabilities, generating precise responses with accurate citations derived specifically from relevant portions of the retrieved context. The color-coding indicates the relevance of retrieved and generated information (green: relevant, red: non-relevant), highlighting Zebra-Llama's improved ability to focus on pertinent information while generating responses.}
\end{figure}

\section{Methods}
In developing Zebra-Llama, our specialized language model for Ehlers-Danlos Syndrome (EDS), we implemented a novel context-aware fine-tuning methodology designed to enhance the model's performance in this specific medical domain. Our approach focuses on equipping the model with high-precision retrieval-augmented generation (RAG) capabilities, aiming to significantly improve its utility in EDS-related queries. The training process integrates diverse data sources, including biomedical literature (PubMed), patient forums (Inspire), and social media discussions (Reddit), which undergo meticulous transformation into structured question-context-answer triplets. This carefully curated dataset is then used to fine-tune the Llama 3 model, with particular emphasis on enhancing its ability to leverage contextual information effectively. Our methodology specifically targets improvements in thoroughness, accuracy, and clarity of responses, while also prioritizing the model's capacity to provide appropriate citations. By focusing on these aspects, we aimed to develop a model capable of generating comprehensive, precise, and well-substantiated information within the EDS domain, thereby addressing the unique challenges of rare disease information retrieval and dissemination.

\subsection{Data Collection}
To develop a comprehensive and nuanced understanding of Ehlers-Danlos Syndrome (EDS), we implemented a sophisticated data collection and preparation process that leverages diverse information sources. Our approach began with the consolidation of data from three primary sources: biomedical literature from PubMed, community discussions on Reddit, and patient experiences shared on the Inspire platform. This multi-faceted dataset formed the foundation for generating a wide spectrum of questions, ranging from formal biomedical inquiries to lifestyle-related queries pertinent to EDS. For each generated question, we extracted relevant context from a custom-built, EDS-focused vector database, comprising over 50,000 indexed entries from PubMed and NCBI gene reviews. This context-rich information was then utilized in conjunction with GPT-4 to craft detailed, accurate answers. To ensure the veracity and relevance of our training data, the generated answers underwent rigorous verification by subject matter experts. Crucially, we enhanced the model's discriminative capabilities by incorporating negative samples-questions unrelated to EDS-sourced from the same platforms. For these out-of-domain queries, we formulated responses that explicitly stated the question's irrelevance to EDS, thereby instilling EDS-specificity into the model's parameters. This comprehensive approach to data collection and curation was designed to equip Zebra-Llama with both depth of knowledge and the ability to accurately delineate EDS-related information from unrelated queries.

\begin{figure}[H]
    \centering
    \includegraphics[width=1\textwidth]{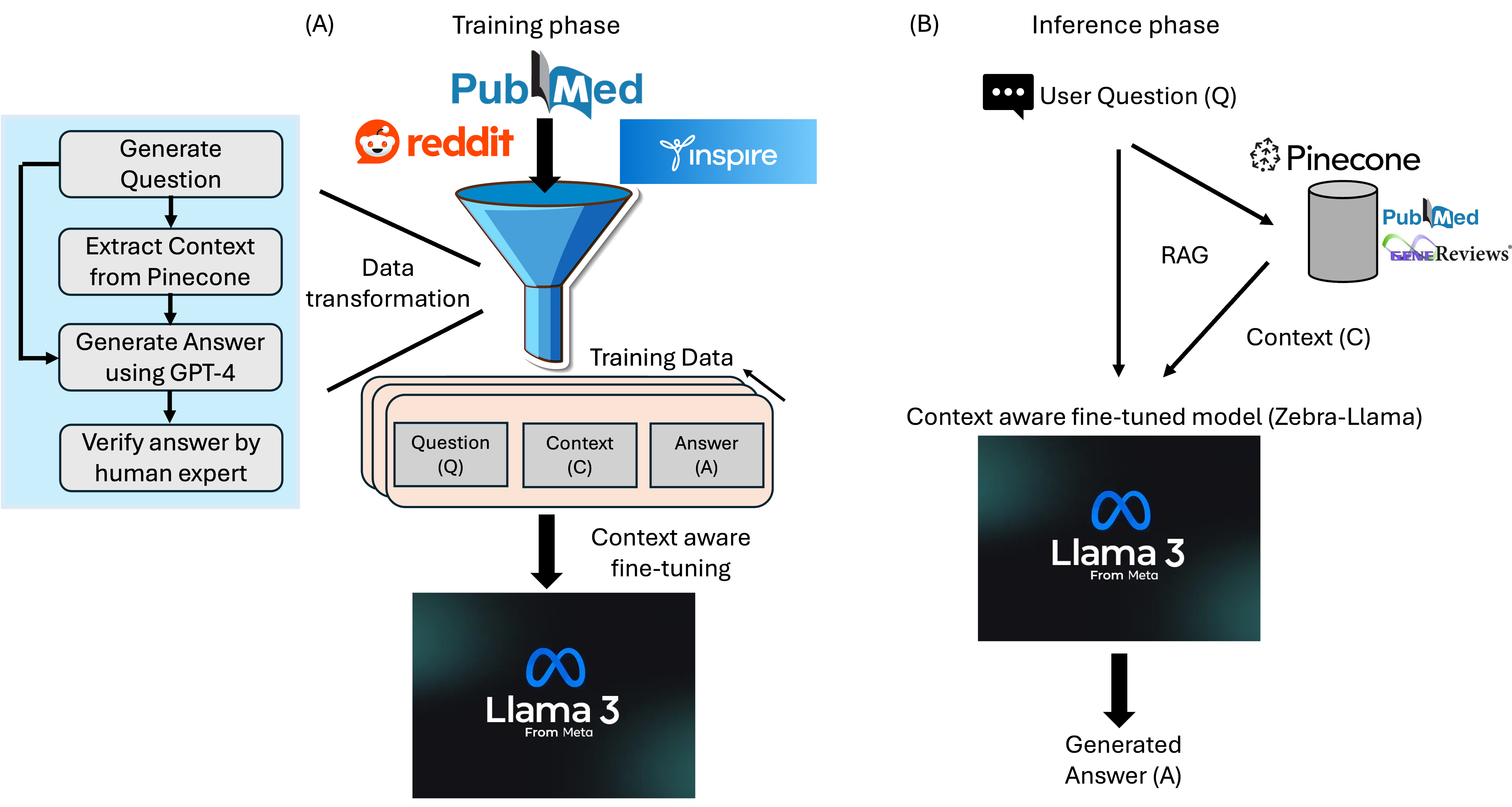}
    \caption{\textbf{Fig 2: Training and Inference Phases of Zebra-Llama} (A) Training Phase: Data from PubMed, Inspire, and Reddit undergoes transformation into structured (Q, C, A) format. The data transformation process is shown in the insight. This processed data is then used for context-aware fine-tuning of Llama-3.1-8B-Instruct model using LoRA. (B) Inference Phase: A user prompt (Q) triggers retrieval of semantically similar documents from the Pinecone vector database, forming the context (C). The fine-tuned Llama 3.1 model then generates the output (A) by processing the concatenated user prompt and retrieved context.}
\end{figure}

\subsubsection{Scientific Literature}
We processed a substantial corpus of 4,371 scientific papers sourced from the National Center for Biotechnology Information (NCBI). This collection comprised:
\begin{itemize}
    \item 1,507 papers in PDF format
    \item 2,864 papers in XML format
\end{itemize}
These academic publications provided a robust foundation of peer-reviewed biomedical knowledge about EDS.

We also processed a corpus of 46 NCBI GeneReviews articles that mention “Ehlers-Danlos” (case insensitive), as of May 9, 2024. The abstract and body were parsed from the XML files, and formatted as text using a Python script. Unlike PubMed scientific papers, GeneReviews articles are designed to provide clinicians, patients, and families with current information on diagnosis and management of inherited conditions, and are kept up-to-date by experts over time. GeneReviews thus provides an excellent overview of the current clinical knowledge on EDS-related conditions.

\subsubsection{Community Discussions: Reddit Posts}
To capture the real-world concerns of the EDS community, we employed a three-pronged approach to collect relevant posts from Reddit:

\paragraph{API-Driven Collection:} Through direct Reddit API calls, we amassed 8,921 unique posts related to EDS, using which we generated a substantial dataset of 8,267 questions. This method allowed us to capture a broad spectrum of EDS-related concerns across various subreddits. The subreddits targeted were ehlersdanlos, Hypermobility, ChronicIllness, Dysautonomia, ChronicPain, POTS, rare\_diseases. The related content contained patient experiences, symptoms, treatments, coping strategies, alternative treatments and healthcare interactions.

\paragraph{Subreddit Scraping:} In addition to API calls, we used the Reddit tool from Langchain to fetch all posts related to EDS from the subreddits "EhlersDanlos," "ChronicPain," and "Hypermobility," resulting in a total of 242 posts. For each post, we generated two unique questions using GPT-3.5-turbo, yielding a total of 484 questions.

\subsubsection{AI-Generated Questions} As a third approach, we generated 602 EDS-related questions using Anthropic's Opus LLM, covering topics such as diet, genetics, lifestyle, and more. Next, we utilized the Reddit Langchain tool to retrieve posts relevant to these questions. Of the 602 questions, only 257 had corresponding posts in Reddit, offering a diverse selection of targeted inquiries. By combining AI-generated questions with related Reddit posts, this approach enabled us to capture only those questions that truly reflect the real-world concerns of the EDS community.

\subsubsection{Patient Support Network: Inspire Online Platform} To further enrich our dataset with questions from patient perspectives, we extracted 5,296 EDS-related posts from the Inspire online community. Inspire is a well-known platform for patients and caregivers to share experiences and support one another, providing valuable insights into the day-to-day realities of living with EDS (\url{https://www.inspire.com/}).

The inclusion of both Reddit and Inspire data was strategic and crucial to our approach. While scientific papers offer formal descriptions and clinical understanding of EDS, these online platforms capture the general concerns that span various dimensions of the disease. By extracting and addressing questions from these discussions, we can better capture:

\begin{itemize}
    \item The everyday challenges faced by individuals with EDS
    \item Common questions and misconceptions about the condition
    \item Coping strategies and lifestyle adjustments
    \item The emotional and psychological aspects of living with a chronic rare disease
    \item Patient-reported symptoms and experiences that may not be fully captured in biomedical literature    
\end{itemize}

Hence, by bridging the gap between formal medical knowledge and patient experiences, we aim to create a more comprehensive AI assistant for the EDS community.

\subsection{Data Transformation}
A key aspect of our approach is the integration of datasets into a unified structured format designed to facilitate context-aware training. For this, we structure our training data in a (Q, C, A) format (Fig 2A):

\begin{itemize}
    \item Q (Question): Question that is extracted or generated from the input data (PubMed, Reddit or Inspire) using different types of LLMs.
    \item C (Context): The context extracted from the relevant document excerpts indexed in the Pinecone vectorDB.
    \item A (Answer): The corresponding answer generated by GPT-4 to the given prompt (Q) + provided context (C). This answer is further vetted to ensure each assertion has proper citations from the provided context. Additionally, a human expert reviews the answer for content accuracy, thereby creating high-quality training data.
\end{itemize}

This structured format, particularly the inclusion of citations within the answer, serves a crucial purpose in model training. By consistently associating assertions with their sources in the training data, we instill in the model a proclivity for providing well-cited responses. This approach enhances the model's ability to ground its generated content in the given context, especially when leveraging retrieval-augmented generation (RAG) during inference. Consequently, Zebra-Llama is trained not only to provide accurate information but also to substantiate its responses with appropriate citations, fostering transparency and traceability in its outputs.

\subsection{Model Training}
We selected the instruction-tuned version of the Llama-3.1 model with 8 billion parameters as our base model. Our fine-tuning process was crafted to strengthen the model's ability to effectively harness the given context from the RAG pipeline. For this, first, we structured our training data in a (Q,C,A) format (Fig. 2A). This tripartite structure mirrors real-world retrieval-augmented generation scenarios, where questions are paired with potentially noisy or variable-quality context. By consistently exposing the model to this format during training, we foster its ability to discern relevant information from the provided context, thereby improving the accuracy and precision in utilizing RAG contexts. This methodology shares similarities with the Retrieval Augmented Fine Tuning (RAFT) technique, as both aim to enhance the model's ability to answer questions in an "open-book" in-domain context \cite{zhang2024raft}. However, our approach differs significantly in its treatment of contextual information. Unlike RAFT, we do not explicitly provide "distractor documents" in the context. Instead, we capitalize on the inherent variability and potential noise in the context retrieved from our vector database. This strategy more closely mimics real-world retrieval scenarios, where the retrieved context can contain a mix of essential and non-essential information, or sometimes no relevant information at all. By training on this naturally occurring noise, our model learns to navigate through imperfect retrievals, developing a robust ability to discern and utilize pertinent information efficiently. This distinction allows our model to adapt to the realities of retrieval-augmented generation in the EDS domain, potentially leading to more accurate and specific responses in practical applications, where the quality and relevance of retrieved information can vary significantly. 

As a next step, we employed a custom system prompt (Appendix A) to set the model's role as an EDS expert assistant, establishing structured response guidelines for addressing user queries with contextual information, proper citations, and maintaining exclusive focus on EDS-related topics.

Next, we fine-tuned the model using the Parameter Efficient Fine-Tuning (PEFT) method, specifically Low-Rank Adaptation (LoRA) \cite{hu2021lora}. Training was conducted on an Amazon Web Service (AWS) p3.16xlarge instance with 8 Tesla V100 GPUs, using the Hugging Face Transformer Reinforcement Learning (TRL) library's Supervised Fine-tuning Trainer. 
(Note: To see the hyperparameters used for training, please check out the YAML file in the GitHub repo of Zebra-Llama \url{https://github.com/karthiksoman/zebra-Llama/blob/main/code/finetuning/model_config.yaml})

\subsection{Retrieval-Augmented Generation (RAG)}
To construct our Retrieval-Augmented Generation (RAG) pipeline, we implemented a comprehensive approach to process and store peer-reviewed scientific literature related to EDS. First, we ingested the full texts of EDS-related papers and NCBI Gene Reviews. We segmented them into chunks of 512 tokens each with an overlap of 128 tokens, striking a balance between granularity and context preservation. Crucially, we enriched these chunks with metadata, including summaries, potential titles, relevant questions, and keywords. This metadata augmentation serves a dual purpose: it captures not only the content but also the context and key insights of each chunk, facilitating more nuanced retrieval and deeper comprehension of the information. To transform these enriched text chunks into a queryable format, we employed OpenAI's text-embedding-ada-002 model, known for its robust performance in capturing semantic relationships. The resulting embeddings were then stored in a Pinecone vector database, creating a rich, searchable knowledge base. This pipeline enables efficient and context-aware retrieval of relevant information, thereby enhancing the model's ability to generate accurate and insightful responses to EDS-related queries.

The vector database we constructed serves a dual purpose in our system. During the training phase, it provides the context (C) component of our (Q,C,A) structured training data, exposing the model to realistic, retrieval-based contexts (Fig. 2A). This same vector database is then leveraged during inference time when users submit queries in real-time (Fig. 2B). By maintaining consistency between the training and deployment environments, we ensure that the model's learned ability to utilize context is directly applicable to real-world scenarios. This approach creates a seamless pipeline where the model's training on retrieved contexts translates effectively to its performance in live query processing. Consequently, when a user submits a query, the system can swiftly retrieve relevant context from the vector database, allowing the model to generate responses that are both accurate and grounded in the most up-to-date EDS-related information available in our corpus.

\subsection{Calibrating EDS Domain Specificity}
To ensure our model's specificity to EDS, we implemented a multi-faceted approach combining system prompt curation, similarity threshold determination, and targeted training strategies. We began by creating a balanced dataset of 200 questions, equally split between EDS-related and non-EDS queries. These questions were generated using GPT-4, guided by carefully crafted prompts to ensure diversity and authenticity in both categories (See Appendix B and C). For non-EDS questions, we focused on distinctly different medical conditions and general topics, while EDS questions reflected varied knowledge levels and common patient concerns (See Appendix B and C).

We then developed a classification mechanism to distinguish between EDS and non-EDS queries. This involved computing embedding vectors for each question using OpenAI's \textit{text-embedding-ada-002} model and querying our EDS-focused vector database. Crucially, for each query, we retrieved the document with the highest cosine similarity score and used this maximum score as the prediction score for that question. The underlying premise was that EDS-related questions would yield higher maximum similarity scores when compared against our specialized database. We framed this as a binary classification problem, with non-EDS and EDS questions labeled as 0 and 1, respectively. To optimize our model's discriminative ability, we computed a precision-recall curve and selected the F2 score as our metric. The F2 score assigns greater weight to recall than precision, thereby prioritizing the minimization of false negatives (missed EDS questions) over false positives. This choice aligns with our goal of ensuring that the model captures as many EDS-related queries as possible, even at the cost of occasionally misclassifying non-EDS questions as relevant. Hence, the similarity score threshold that maximized the F2 score was chosen as our decision boundary.

Our commitment to domain specificity extended to the training phase, where we incorporated negative samples (non-EDS data) to teach the model to refrain from answering out-of-domain questions. Additionally, we augmented the system prompt with explicit instructions and examples of non-EDS questions to reinforce this behavior. This three-pronged approach of curated system prompt, optimized similarity threshold and targeted training, collectively equipped our model to maintain a sharp focus on EDS-related queries while appropriately deflecting unrelated questions, ensuring a high degree of domain specificity in its responses.

\subsection{Model Evaluation}
Our evaluation process for Zebra-Llama was rigorous and multi-faceted, combining expert manual analysis with automated assessment. We utilized a dataset of 51 real-world questions collected from EDS patients and clinicians during a rare disease AI hackathon conducted by Stanford Medicine and Research to the People. This ensured our evaluation was grounded in genuine concerns and queries from the EDS community.

We conducted a comparative analysis between Zebra-Llama and base-Llama, both provided with RAG context (with top-k=2) for fair comparison. The generated answers were anonymized to prevent bias. The responses were evaluated by three expert evaluators (two medical doctors who conduct clinical research at University of California San Francisco (UCSF) and one non-clinical subject matter expert on EDS), with each evaluator assessing a portion of the test questions to collectively cover the entire test dataset. The expert evaluators assessed the responses based on three criteria: Thoroughness (0-100), Accuracy (0-100), and Clarity (0-100). These criteria were designed to comprehensively evaluate the quality, correctness, and comprehensibility of the generated answers. To validate the manual analysis, we performed an independent automated evaluation using GPT-4, applying the same criteria. We then computed the inter-rater intra-class correlation coefficient \cite{bartko1966intraclass} to measure the agreement between the manual and automated assessments, providing an additional layer of validation to our evaluation process.

Furthermore, we conducted a citation accuracy analysis for both Zebra-Llama and base-Llama. This involved extracting citation links (DOIs) from the generated text and verifying their validity. This analysis offered insights into the models' ability to provide correct and verifiable citations, a crucial aspect in the medical domain.

This comprehensive evaluation approach, combining expert human judgment, automated assessment, and citation verification, allowed us to thoroughly assess Zebra-Llama's performance in addressing real-world EDS queries, ensuring a robust and multidimensional evaluation of our model's capabilities.

\section{Results}

\subsection{EDS Domain Specificity}
Fig. 3 illustrates the efficacy of our EDS domain specificity mechanism. Fig. 3A depicts the distribution of similarity scores for EDS and non-EDS questions, revealing distinct semantic patterns. EDS questions cluster tightly around 0.85 ± 0.02, while non-EDS questions show a broader distribution centered at 0.79 ± 0.05. Notably, the overlapping region between these distributions corresponds to medical-related questions present in both datasets, reflecting real-world complexity where some queries may share characteristics across domains. For instance, EDS-related questions like "Can you still exercise if you have EDS?" and non-EDS medical queries such as "What are the treatment options for post-traumatic stress disorder (PTSD)?" fall within this overlap, challenging the model's discriminative capabilities. This nuanced separation underscores the model's ability to differentiate EDS-related queries while acknowledging the challenges posed by other related medical topics.

Fig. 3B presents the precision-recall curve for classifying these question types. We identified an optimal threshold of 0.81, maximizing the F2 score with a high recall of 0.98 and precision of 0.74. This threshold, favoring recall over precision, minimizes false negatives which is crucial for ensuring EDS-related queries are not overlooked. During inference, our RAG system utilizes this threshold to determine context provision, only retrieving information for questions exceeding this similarity score.

The practical application of this approach is exemplified in Fig. 3A, where Zebra-Llama demonstrates its discriminative capabilities. It provides a detailed, citation-backed response to an EDS-specific query about hEDS mutations, while appropriately declining to answer an unrelated question about “how many days are there in a week”.

\begin{figure}[H]
    \centering
    \includegraphics[width=1\textwidth]{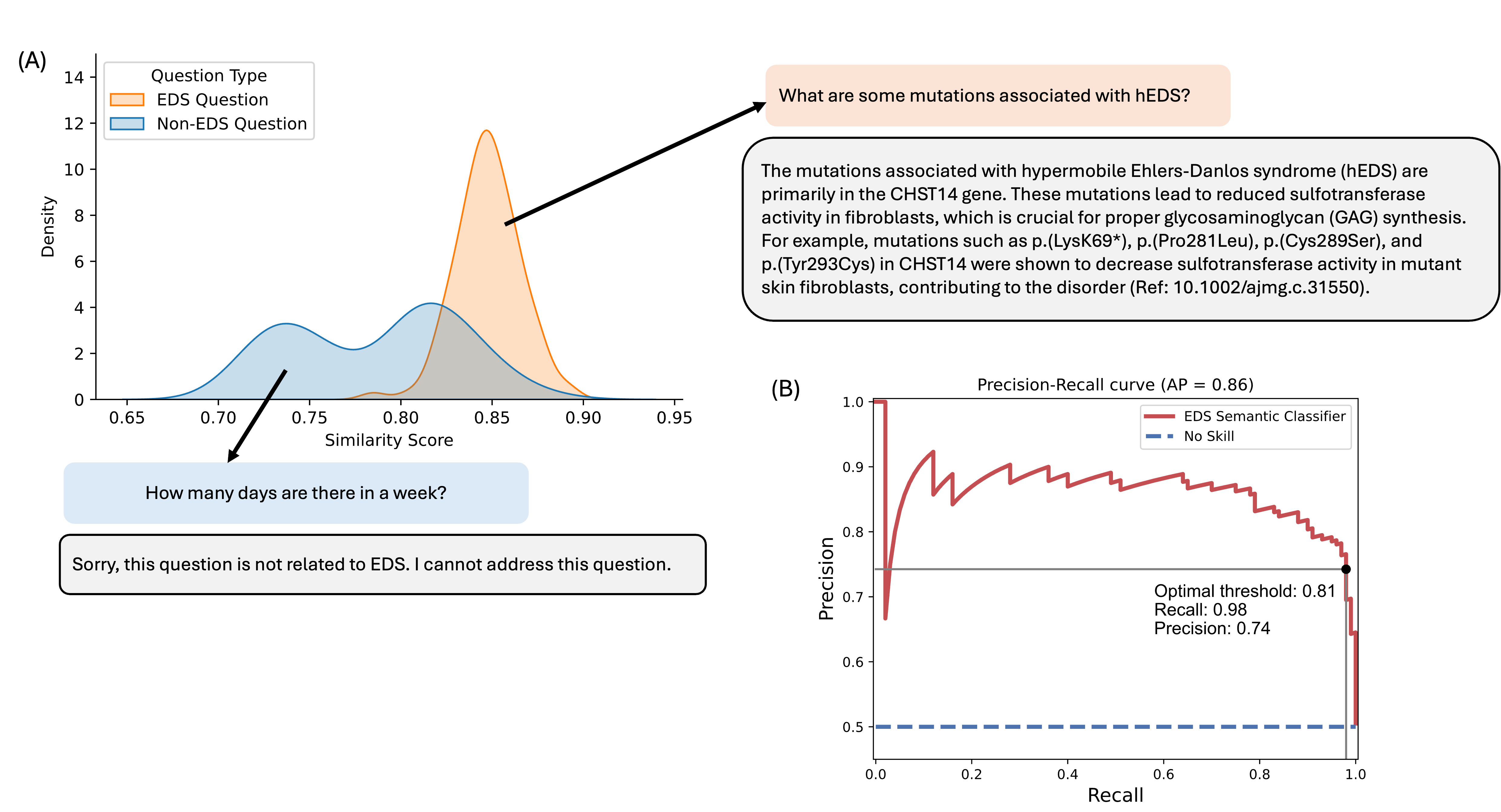}
    \caption{\textbf{Fig 3: EDS domain specificity evaluation through similarity score distribution and classification performance} (A) Distribution of similarity scores for EDS-related (orange) and non-EDS (blue) questions, demonstrating distinct semantic patterns with example queries and responses. EDS questions cluster around higher similarity scores (0.85 ± 0.02), while non-EDS questions show a broader distribution (0.79 ± 0.05). (B) Precision-Recall curve for the EDS semantic classifier, achieving an optimal threshold of 0.81 with high recall (0.98) and precision (0.74). The classifier substantially outperforms the no-skill baseline (AP = 0.86), indicating robust discrimination between EDS and non-EDS queries.}
\end{figure}

\subsection{Model Evaluation}
Our rigorous evaluation compared Zebra-Llama and base-Llama on a dataset of 51 real-world EDS questions collected from EDS patients and clinicians (See Methods), with both models provided RAG context. This approach allowed us to assess not only the models' general performance but also their specific abilities to effectively utilize retrieved information in generating responses with proper citations.

Fig. 4A illustrates the manual expert analysis on the comparative performance of Zebra-Llama and base-Llama on three key metrics: thoroughness, accuracy, and clarity. Zebra-Llama consistently outperformed base-Llama across all criteria. Notably, Zebra-Llama achieved an average thoroughness score of 77.5\% compared to base-Llama's 70.1\%, demonstrating its enhanced ability to provide comprehensive responses. In terms of accuracy, Zebra-Llama scored 83.0\%, surpassing base-Llama's 78.8\%, indicating a marked improvement in the correctness of information provided. Clarity scores also favored Zebra-Llama (74.7\% vs. 72.0\%), suggesting that our model's responses were more comprehensible and well-structured.

To validate the robustness of our manual evaluation, we conducted an automated assessment using GPT-4 and compared it with the expert evaluations. Fig. 4B presents the correlation between manual and automated assessments for each criterion. The inter-rater intraclass correlation coefficients (ICC) were 0.675 for thoroughness, 0.576 for accuracy, and 0.608 for clarity which indicate moderate agreement between human experts and the automated system, lending credibility to our evaluation methodology.

A critical aspect of medical AI systems is their ability to provide accurate citations, and Zebra-Llama demonstrated significant improvements in this area. Fig. 4C illustrates the per-response citation accuracy, measuring the percentage of correct citations within each response. Zebra-Llama demonstrated superior performance, with an average per-response citation accuracy of 70.4\%, significantly outperforming base-Llama's 52.3\%. This metric reflects Zebra-Llama's enhanced capability to provide a higher proportion of accurate citations within individual responses. Complementing this, Fig. 4D shows the percentage of responses where all citations are correct. Zebra-Llama again excelled, with 68.2\% of its responses containing only correct citations, compared to base-Llama's 51.4\%. This metric highlights Zebra-Llama's improved consistency in providing entirely accurate citation sets across different queries. We further investigated if the higher citation performance shown by Zebra-Llama could be attributed to any data leakage between training and test RAG contexts or whether it demonstrated true learning of context utilization. For this evaluation, we generated 31 EDS questions and ensured their RAG contexts were entirely unseen during the training phase. Even with completely novel contexts, Zebra-Llama maintained its superior performance, achieving a higher per-response citation accuracy (82.1\% vs 75.0\%) and outperforming base-Llama in the percentage of responses with all correct citations (78.6\% vs 64.3\%) (Appendix D).

\begin{figure}[H]
    \centering
    \includegraphics[width=1\textwidth]{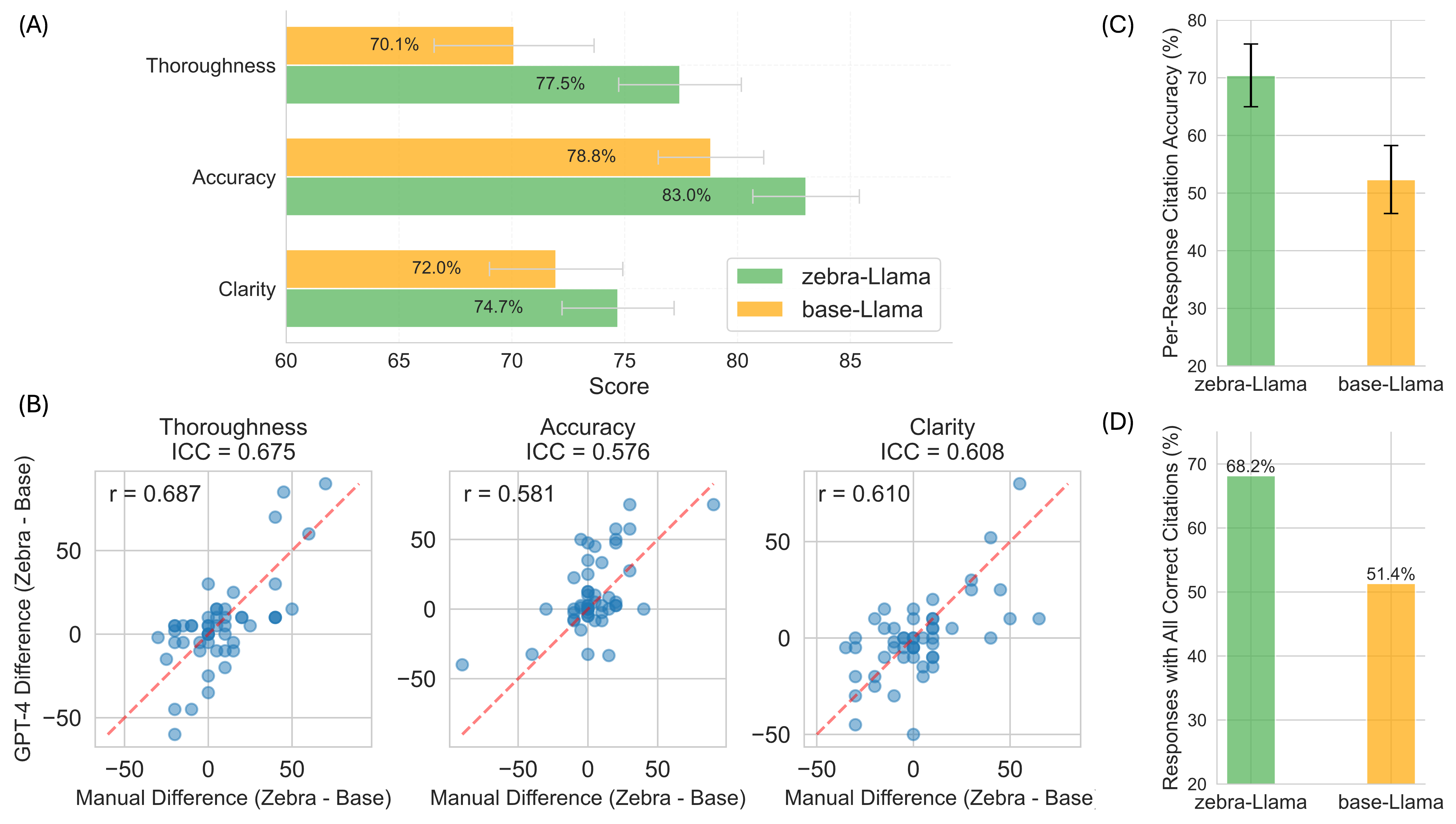}
    \caption{\textbf{Fig 4: Comprehensive evaluation of Zebra-Llama's performance} (A) Expert manual evaluation comparing performance metrics between Zebra-Llama and base-Llama, showing improvements in thoroughness (77.5\% vs 70.1\%), accuracy (83.0\% vs 78.8\%), and clarity (74.7\% vs 72.0\%). Error bars indicate s.e.m (B) Correlation analysis between manual expert evaluation and automated GPT-4 assessment, demonstrating moderate agreement (agreement is quantified using Intraclass Correlation Coefficient-ICC and "r" denotes Pearson correlation coefficient) across all metrics (Thoroughness: ICC = 0.675, r = 0.687; Accuracy: ICC = 0.576, r = 0.581; Clarity: ICC = 0.608, r = 0.610). (C) Per-response citation average accuracy comparison, showing Zebra-Llama's superior performance (70.4\% ± 5.4\%) compared to base-Llama (52.3\% ± 5.9\%). Error bars indicate s.e.m (D) Percentage of responses with all correct citations, with Zebra-Llama (68.2\%) outperforming base-Llama (51.4\%), indicating improved citation reliability.}
\end{figure}

\section{Discussion}
In this study, we introduced Zebra-Llama, a specialized large language model fine-tuned for the domain of EDS and released as an open-source resource. Our approach addressed the critical need for accurate, comprehensive, and verifiable information in the rare disease space, particularly for EDS. By implementing a novel context-aware fine-tuning methodology, we enhanced the base Llama model's ability to effectively utilize retrieved information and maintain a sharp focus on EDS-related queries. The resulting Zebra-Llama model demonstrated significant improvements across multiple dimensions, including thoroughness, accuracy, clarity of responses, and citation reliability. Our rigorous evaluation, combining expert manual analysis with automated assessment, validated Zebra-Llama's superior performance in handling real-world EDS queries. This specialized model not only outperformed its base counterpart in generating relevant and accurate responses but also showed marked improvement in providing verifiable citations, a crucial aspect for trustworthy medical information systems. The development and open-source release of Zebra-Llama represents a significant step forward in leveraging AI to address the unique challenges faced by the EDS community, potentially improving access to reliable information for patients, caregivers, and healthcare professionals alike while fostering collaborative advancement in rare disease informatics.

Our context-aware fine-tuning methodology introduces several key innovations in enhancing LLMs for domain-specific applications. While our method shares some conceptual similarities with Retrieval Augmented Fine Tuning (RAFT) \cite{zhang2024raft}, it differs in several key aspects. Unlike RAFT, which explicitly introduces distractor documents during training, our approach leverages the natural variability in retrieved information from an EDS-focused vector database. This design choice eliminates the need for artificial distractor generation, instead training the model on realistic information retrieval scenarios. The core of our algorithm lies in the structured (Q,C,A) format, where Q represents the query, C the retrieved context, and A the answer. Crucially, we incorporate a dynamic context retrieval process during training, where the relevance of retrieved information varies naturally. This approach trains the model to adaptively focus on pertinent information within noisy contexts. The model is also trained to generate answers (A) that include explicit citations derived from the context (C), fostering a direct link between retrieved information and model outputs. Due to this emphasis on correct citations from the context during training, the model inherently develops the ability to extract relevant information from inherently noisy RAG contexts (“needle in the haystack” problem \cite{nelson2024needlehaystackmemorybased, liu2023lostmiddlelanguagemodels}).

The calibration of Zebra-Llama's domain specificity revealed intriguing insights into the challenges and opportunities of developing highly specialized medical AI systems. Our approach demonstrated that achieving a balance between sensitivity to in-domain queries and rejection of out-of-domain questions is crucial yet complex. The distribution of similarity scores between EDS and non-EDS questions highlighted a nuanced overlap, particularly in general medical topics, reflecting the interconnected nature of medical knowledge. This overlap underscores the importance of fine-grained discrimination in medical AI, where the boundaries between specialties can be subtle. The prioritization of recall over precision in our F2 score optimization reflects a strategic choice in medical AI development: it's often preferable to cast a slightly wider net in information retrieval than to miss critical, relevant information. This approach ensures comprehensive coverage of EDS-related topics, which is vital in a rare disease context where information scarcity is a significant challenge. Furthermore, the incorporation of negative samples in training proved essential in developing the model's discernment capabilities, mimicking the real-world scenario where healthcare professionals must constantly distinguish between relevant and irrelevant information. These insights not only guided the development of Zebra-Llama but also offer valuable lessons for the broader field of specialized medical AI, highlighting the need for nuanced, domain-aware training strategies that can navigate the complex landscape of medical knowledge.

The comprehensive evaluation of Zebra-Llama offers valuable insights into the potential of specialized AI models in the rare disease domain. The model's superior performance in thoroughness, accuracy, and clarity compared to the base model underscores the transformative impact of our context-aware, domain-specific fine-tuning approach. This methodology proves particularly crucial in the rare disease context, where information is often scattered, complex, and highly contextual \cite{griggs2009clinical, stoller2018challenge, nestler2018challenges}. Our method deliberately focuses on enhancing the model's ability to effectively utilize available contexts with high precision. This focused approach is especially relevant for rare conditions like EDS, where extracting maximum value from each available piece of information is paramount. By training the model to excel at precise interpretation and citation of available context, we address a fundamental challenge in rare disease knowledge management: the need to make optimal use of limited but crucial information sources.

The significant enhancement in citation accuracy further highlights a critical advancement for rare disease information systems. In a field where research is limited and rapidly evolving, the ability to not only provide accurate information but also ground it in verifiable, context-appropriate sources is paramount for building trust among patients and healthcare professionals. This also addresses the need for reliable, up-to-date, and contextually relevant information to guide clinical decisions and patient care. Moreover, the model's ability to maintain high performance across a diverse range of real world EDS concerns, while effectively leveraging provided context, demonstrates its potential to serve as a dynamic, comprehensive knowledge resource. Collectively, these results indicate that context-aware AI, when carefully tailored to specific rare diseases, can play a transformative role in democratizing access to expert-level, contextually relevant knowledge, potentially improving diagnosis rates, treatment strategies, and overall patient care in fields where expertise is scarce and information needs are complex and highly contextual.

The open-source release of Zebra-Llama represents a significant step towards democratizing advanced AI technologies in the medical domain, particularly for rare disease research. By making our context-aware, EDS-specialized model freely available to the global research community, we aim to catalyze collaborative efforts and accelerate progress in understanding and managing Ehlers-Danlos Syndrome. This open approach is especially crucial in the rare disease landscape, where resources are often limited and collaboration is key to advancing knowledge. The availability of Zebra-Llama as an open-source tool can potentially level the playing field, allowing researchers, clinicians, and even patient advocacy groups from diverse backgrounds and resource settings to access state-of-the-art AI capabilities tailored to EDS. Furthermore, the transparent nature of open-source projects fosters trust and enables continuous improvement through community contributions. This model can serve as a foundational resource, spurring the development of similar specialized models for other rare diseases, ultimately creating a network of AI tools that address the unique challenges of various understudied conditions. Hence, by embracing open source in the medical AI domain, we not only contribute to the democratization of technology but also pave the way for a more inclusive, collaborative, and rapidly evolving landscape of rare disease research and management.

While Zebra-Llama represents a significant advancement in AI-assisted EDS information management, we acknowledge several limitations and areas for future development. The model's current knowledge is constrained by the available EDS literature and may not fully capture the most recent research or rare manifestations of the syndrome. Developing a robust update mechanism to keep the model current with emerging EDS research is a critical next step. Future work should also focus on enhancing the model's explainability, allowing users to better understand the reasoning behind its responses. Integration of Zebra-Llama into clinical workflows presents both opportunities and challenges that warrant further investigation, particularly in ensuring ethical use and maintaining patient privacy. Looking ahead, we envision applying our context-aware fine-tuning methodology to other rare diseases, potentially creating a network of specialized AI models to support underserved medical communities. Longitudinal studies assessing the long-term impact of such AI tools on patient outcomes and clinical practice will be crucial in validating and refining their role in rare disease management. As we continue to develop and refine Zebra-Llama, we remain committed to addressing these challenges and exploring new avenues to maximize its potential in improving EDS care and research.

In conclusion, Zebra-Llama demonstrates the potential of context-aware, specialized AI in addressing the unique challenges of rare disease information management. By significantly improving thoroughness, accuracy, clarity, and citation reliability in EDS-related queries, our open-source model paves the way for more accessible and reliable rare disease knowledge. This advancement marks a crucial step towards democratizing expert-level information for EDS, potentially enhancing clinical decision-making and patient care. Moreover, it sets a precedent for developing similar AI-driven solutions for other rare diseases.

\section{Author contributions}
KS conceived and led the project, architected the Zebra-Llama model, developed the context-aware fine-tuning methodology, implemented the data processing and model fine-tuning, conducted model evaluation and downstream analyses, spearheaded project discussions and wrote the manuscript. AL contributed to training data generation and vector database development, including document processing and indexing, participated in model fine-tuning, contributed to downstream analyses and engaged in project discussions. CV contributed to data curation and training dataset development, maintained project documentation, designed the project logo, and actively participated in model evaluation and strategic discussions. CA contributed to data curation, transformation and training dataset development, and developed the serverless infrastructure for model deployment and inference while contributing to technical discussions. LS served as EDS subject matter expert, contributed to training dataset generation and model evaluation, and provided valuable input through project discussions. BP contributed to model evaluation, offered important clinical guidance for the project, and participated in project discussions. GB contributed to model evaluation and manuscript preparation, provided valuable clinical insights, and shared perspectives during project discussions. OJB contributed to dataset generation, provided methodological feedback, supported project promotion, and participated in manuscript review.

\section{Acknowledgment}
The authors sincerely acknowledge "Research to the People" and Stanford Medicine for providing a valuable corpus of EDS-related data as part of the rare disease AI hackathon \url{https://rttp.stanford.edu/events}. We also thank the Inspire platform for granting access to anonymized conversation data from the EDS community. Special thanks to David Harris, patient advocate and practice manager at "The EDS Clinic" for his support and productive guidance on EDS-related topics throughout. We appreciate Amazon Web Services (AWS) for providing cloud computing credits that enabled model training and downstream analyses, and OpenAI for granting API credits to the corresponding author (KS) through their "Researcher Access Program".

\bibliographystyle{unsrt}  
\bibliography{references}

\newpage
\appendix
\section{Appendix A: Zebra-Llama System Prompt}
\ttfamily
You are an expert AI assistant specializing in Ehlers-Danlos syndrome (EDS). Your role is to provide comprehensive, accurate, and well-structured answers about EDS. You will be provided with a prompt that has two components such as "User message" and "Context". Follow these guidelines to address the prompt:

- In the first paragraph, begin with a broad overview that directly addresses the "User message".

- In the second paragraph, provide detailed information mainly by using the given "Context". Also use your trained knowledge about EDS to supplement the assertions. If you don't see relevant information in the context, always mention that in your response and stick on to your own internal knowledge to answer the question.

- Answer in multiple paragraphs and be comprehensive in your answer

- Structure your response logically:

     a) Start with a general answer to the question.
     
     b) Provide specific examples or details, always with proper citations. 
     
     c) You can find the citations at the end of each "Context" para marked as '(Ref: '. Do not use any references that do not contain a DOI, and do not use references that contain just numbers in square brackets. Here are examples of references to avoid: [ 1 ], [5, 6, 8], etc.
     
- If mentioning specific studies or cases, clearly state their relevance to the main question and provide proper context.

- In the last paragraph, conclude with a brief summary of the key points.

IMPORTANT: If you receive a question unrelated to Ehlers-Danlos Syndrome (EDS), respond directly by stating that the question is not related, without providing any additional context or explanations. For example, if the question is "Who is the actor in the movie titanic" and even if it has any EDS context given in the "Context", your answer should be like "Sorry, this question is not related to EDS and I cannot address that."

\newpage
\normalfont
\section{Appendix B: Prompt for generating synthetic EDS questions for domain specificity analysis}
\ttfamily
You are an AI assistant specialized in Ehlers-Danlos Syndrome (EDS). Your task is to generate 100 realistic questions that a person might ask about EDS. These questions should reflect the following characteristics:

1. Varied knowledge levels: Generate questions from people with different levels of understanding about EDS, from newly diagnosed patients to those who have lived with it for years.

2. Common concerns: Focus on typical issues EDS patients face, such as joint hypermobility, skin elasticity, chronic pain, fatigue, and related complications.

3. Realistic language: Use natural language patterns that people typically use when asking questions. This includes:

   - Occasional typos or misspellings

   - Use of acronyms (e.g., EDS, POTS, MCAS)

   - Informal language or slang terms

   - Incomplete sentences or fragmented thoughts
   
4. Question types: Include a mix of question types, such as:

   - Seeking general information about EDS

   - Asking about specific symptoms or complications

   - Inquiring about treatment options or pain management

   - Requesting advice on daily living with EDS

   - Asking about genetic aspects or inheritance patterns

5. Emotional context: Some questions may reflect the emotional challenges of living with EDS, such as frustration, worry, or hope for new treatments.

Generate 100 diverse questions that a real person might ask about EDS, incorporating these elements to create authentic, varied queries.

Please respond with the following format: 

[
{"instruction": <question goes here>}
]

\newpage
\normalfont
\section{Appendix C: Prompt for generating synthetic non-EDS questions for domain specificity analysis}
\ttfamily

You are an AI assistant tasked with generating diverse questions that are NOT related to Ehlers-Danlos Syndrome (EDS). Your goal is to create a variety of questions that a general-purpose AI might encounter, ensuring they are distinctly separate from EDS-related topics. Generate 100 questions following these guidelines:

1. Other medical conditions (40\% of questions): Create questions about diseases or health issues that are not related to EDS or its common comorbidities. Focus on conditions with symptoms, treatments, or characteristics clearly different from EDS.

   Examples:
   
   - "What are the early signs of Parkinson's disease?"

   - "How is type 2 diabetes typically managed?"

   - "What causes seasonal allergies and how can they be treated?"

   - "What are the risk factors for developing osteoporosis?"
   
2. General health and wellness (20\% of questions): Include questions about general health practices, nutrition, or fitness that don't overlap with EDS management.

   Examples:
   
   - "How many hours of sleep does an average adult need?"

   - "What are the benefits of a Mediterranean diet?"

3. General knowledge (10\%): Include questions about history, geography, science, or current events.

   Example: "What is the capital of France?"
   
4. Entertainment and pop culture (10\%): Generate questions about movies, music, books, or celebrities.

   Example: "Who played the lead role in the latest Marvel movie?"

5. Technology and gadgets (10\%): Include questions about computers, smartphones, or other tech topics.

   Example: "How do I set up two-factor authentication on my email account?"
   
6. Everyday life and practical matters (10\%): Generate questions about cooking, gardening, home maintenance, or other daily activities.

   Example: "What's the best way to remove a red wine stain from a carpet?"
   
Important guidelines:

- Ensure that none of the questions are related to EDS, its symptoms, or commonly associated conditions (like POTS, MCAS, or joint hypermobility).

- Avoid questions about connective tissue disorders, chronic pain, or genetic disorders that might be confused with EDS.

- Do not include questions about flexibility, skin elasticity, or joint problems, as these could be mistaken for EDS-related queries.

- When creating health-related questions, focus on conditions with distinct symptoms, causes, or treatments from EDS (e.g., infectious diseases, metabolic disorders, neurological conditions).

- Maintain diversity in the types of questions to represent a wide range of topics.

Generate 100 diverse questions that are clearly unrelated to EDS, incorporating these elements to create a set of negative samples for training purposes. Ensure that at least 40\% of these questions are about specific medical conditions or health issues that are distinctly different from EDS.

use this format:

[
{"instruction": <question goes here>}
]

\newpage
\normalfont
\section{Appendix D: Validation of Citation Performance on unseen RAG Contexts}

\begin{figure}[H]
    \centering
    \includegraphics[width=1\textwidth]{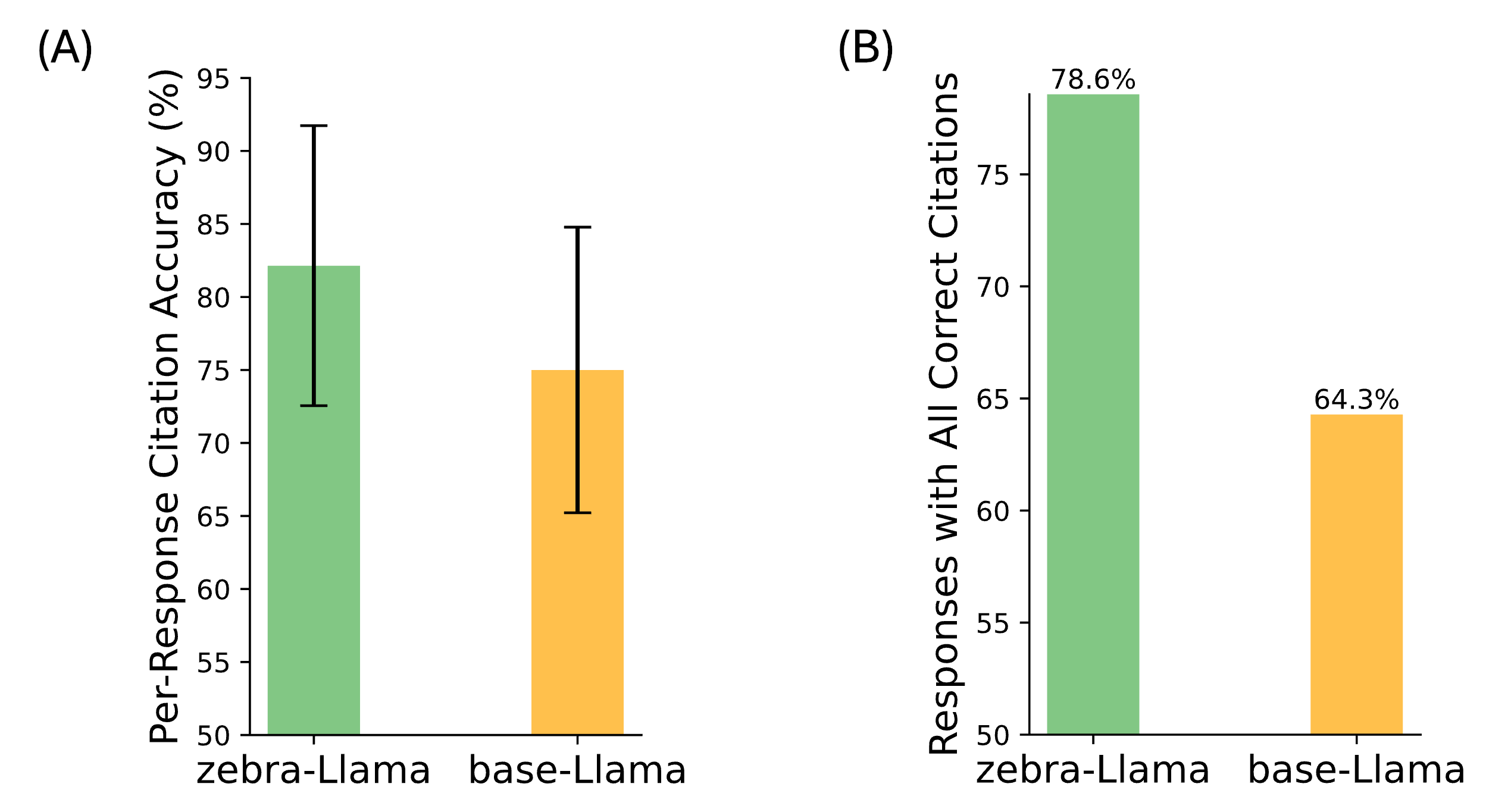}
    \caption{\textbf{Citation performance validation on unseen RAG contexts} (A) Per-response citation accuracy comparison between Zebra-Llama (82.1\% ± 9.6\%) and base-Llama (75.0\% ± 9.8\%) on test questions with entirely unseen contexts (metric is given as mean ± sem) (B) Percentage of responses with all citations correct, showing Zebra-Llama (78.6\%) maintaining superior performance over base-Llama (64.3\%) when evaluated on novel contexts. These results validate that Zebra-Llama's enhanced citation capabilities persist even when handling previously unseen RAG context.}
\end{figure}

\end{document}